\documentclass{article}

\usepackage{microtype}
\usepackage{graphicx}
\usepackage{subcaption}
\usepackage{booktabs}
\usepackage{hyperref}

\usepackage[accepted]{icml2026}

\usepackage{amsmath}
\usepackage{amssymb}
\usepackage{amsfonts}
\usepackage{nicefrac}
\usepackage{xcolor}
\usepackage{placeins}
\usepackage{xurl}

\icmltitlerunning{Stable and Steerable SAEs with Weight Regularization}

\begin{document}

\twocolumn[
  \icmltitle{Stable and Steerable Sparse Autoencoders\\with Weight Regularization}

  \icmlsetsymbol{equal}{*} 
  \icmlsetsymbol{corres}{$\dagger$}

  \begin{icmlauthorlist}
    \icmlauthor{Piotr Jedryszek}{bio,corres}
    \icmlauthor{Oliver M. Crook}{kavli,chem}
  \end{icmlauthorlist}

  \icmlaffiliation{bio}{Department of Biology, University of Oxford, Oxford, UK}
  \icmlaffiliation{kavli}{Kavli Institute for Nanoscience Discovery, University of Oxford, Oxford, UK}
  \icmlaffiliation{chem}{Department of Chemistry, University of Oxford, Oxford, UK}

  \icmlcorrespondingauthor{Piotr Jedryszek}{piotr.jedryszek@sjc.ox.ac.uk }

  \icmlkeywords{Sparse Autoencoders, Mechanistic Interpretability, Weight Regularization, Feature Stability}

  \vskip 0.3in
]

\printAffiliationsAndNotice{}

\begin{abstract}
Sparse autoencoders (SAEs) are widely used to extract human-interpretable features from neural network activations, but their learned features can vary substantially across random seeds and training choices. To improve stability, we studied \emph{weight regularization} by adding L1 or L2 penalties on encoder and decoder weights, and evaluate how regularization interacts with common SAE training defaults. On MNIST, we observe that L2 weight regularization produces a core of highly aligned features and, when combined with tied initialization and unit-norm decoder constraints, it dramatically increases cross-seed feature consistency. For TopK SAEs trained on language model activations (Pythia-70M-deduped), adding a small L2 weight penalty increased the fraction of features shared across three random seeds and roughly doubles steering success rates, while leaving the mean of automated interpretability scores essentially unchanged. Finally, in the regularized setting, activation steering success becomes better predicted by auto-interpretability scores, suggesting that regularization can align text-based feature explanations with functional controllability. Code including for training and analysis is available at \url{https://github.com/oxPJ/sae-weight-regularization}, trained SAE's and result data is available at \url{https://huggingface.co/anonsaereg/SAE-REG-models} and \url{https://huggingface.co/datasets/anonsaereg/SAE-REG-results} respectively. 
\end{abstract}

\section{Introduction}

Sparse autoencoders (SAEs) have become central to mechanistic interpretability, offering a potential solution to the superposition hypothesis, the idea that neural networks represent more features than dimensions by encoding them in overlapping patterns \citep{Elhage2022ToySuperposition}. By learning an overcomplete basis with sparse coefficients, SAEs aim to recover the ``true'' features underlying model computations \citep{Bricken2023TowardsLearning, Cunningham2023SparseModels}. Despite their success, recent work raised concerns about SAE reliability. \citet{Chanin2025SparseAutoencoders} showed that SAE features are highly sensitive to the L0 sparsity hyperparameter, and feature splitting across dictionary sizes \citep{Bricken2023TowardsLearning} similarly demonstrates that which features are learned depends on model capacity. \citet{Paulo2025SparseFeatures} demonstrated that SAEs trained with different random seeds on identical data learn substantially different features. This variability suggests \textbf{underconstrained optimization}: activation sparsity alone does not uniquely determine a solution.

This underdetermination is consistent with mixed downstream results: in a large-scale study of activation probing, \citet{Kantamneni2025AreProbing} find that SAE-based probes do not provide a consistent advantage over strong baselines on raw activations across several challenging regimes, although they sometimes outperform on individual datasets. Recent work also argues that SAE objectives should prioritize \emph{functional faithfulness} to model behavior rather than reconstruction alone; that is, by training SAEs end-to-end to preserve model outputs \citep{Braun2024IdentifyingLearning}. Other related work has begun to address these challenges by adding additional structure to SAE training beyond activation sparsity. For example, \citet{Marks2024EnhancingAutoencoders} propose \emph{Mutual Feature Regularization}, which trains multiple SAEs in parallel and encourages them to converge to similar features, motivated by the idea that features replicated across runs are more likely to correspond to genuine input factors. Similarly, recently \citet{Martin-Linares2025Attribution-GuidedAutoencoders} 
introduced Distilled Matryoshka Sparse Autoencoders (DMSAEs), which run iterative 
train-and-select cycles: after each training run, features are scored by their 
gradient$\times$activation contribution to next-token loss, and only the 
highest-attribution subset is carried forward as frozen encoder directions for 
the next cycle. Where Mutual Feature Regularization encourages convergence across 
\emph{parallel} SAEs, DMSAEs enforce consistency across \emph{sequential} 
retraining cycles.

In classical machine learning, regularization is often used to bias underdetermined or overparameterized optimization problems toward simpler solutions by penalizing particular classes of parameters \citep{Goodfellow2016DeepLearning}. L1 regularization (Laplace prior) encourages weight sparsity, while L2 regularization (Gaussian prior) encourages small, smoothly distributed weights. Both improve generalization in many settings \citep{Tibshirani1996RegressionLasso, Krizhevsky2012ImageNetNetworks}. In an overcomplete setting, however, weight penalties can also affect the effective size of the learned dictionary by suppressing weakly-used features — a mechanism we engage with directly when interpreting our results.

In this work we revisit a simple idea: add an explicit weight penalty to SAE training, in addition to the usual activation sparsity term. We focus on three questions:
\begin{enumerate}
  \item \textbf{Cross-seed consistency:} Does weight regularization increase the reproducibility of features across random seeds?
  \item \textbf{Feature quality:} Does weight regularization affect downstream evaluations such as interpretability and steering?
  \item \textbf{Interaction with SAE design choices:} How does weight regularization interact with common architectural and training decisions such as encoder/decoder initialization, decoder constraints, sparsity mechanism (TopK, BatchTopK, Matryoshka)? 
\end{enumerate}

We explore weight regularization first on a toy model of MNIST images to build intuitions, then on a small language model (Pythia-70M-deduped) to test real-world applicability. 

\section{Methods}

\subsection{Sparse autoencoders with weight regularization}

Given an activation vector $x \in \mathbb{R}^{d}$, an SAE produces a sparse latent $z \in \mathbb{R}^{m}$ and a reconstruction $\hat{x}$:
\begin{align}
z &= f(W_{\mathrm{enc}} x + b_{\mathrm{enc}}), \\
\hat{x} &= W_{\mathrm{dec}} z + b_{\mathrm{dec}}.
\end{align}
We consider standard sparsity mechanisms for $z$ (e.g.\ an explicit $\ell_1$ penalty or a hard TopK constraint) and add weight regularization:

\begin{equation}
\begin{split}
\mathcal{L} = {}& \mathcal{L}_{\mathrm{recon}}
  + \lambda_{\mathrm{sparse}}\, \mathcal{L}_{\mathrm{sparse}}(z) \\
  &+ \lambda_{\mathrm{w}} \Big( \|W_{\mathrm{enc}}\|_{p}^{p}
  + \|W_{\mathrm{dec}}\|_{p}^{p} \Big),
\end{split}
\end{equation}
where $p\in\{1,2\}$ and $\mathcal{L}_{\mathrm{recon}} = \|x - \hat{x}\|_2^2$.
For TopK models, sparsity is enforced by the encoder and we set $\mathcal{L}_{\mathrm{sparse}}(z)=0$.

\subsection{Decoder constraints and tied initialization}

Two implementation choices are common in recent SAE work and are present in the default SAE implementations of SAEBench \citep{Karvonen2025SAEBench:Interpretability}:
\begin{enumerate}
  \item \textbf{Tied initialization:} initialize $W_{\mathrm{dec}} \leftarrow W_{\mathrm{enc}}^{\top}$ so that each latent starts as an approximately self-consistent ``detector'' and ``writer'' direction.
  \item \textbf{Unit-norm decoder columns:} enforce $\|W_{\mathrm{dec}}(:,i)\|_2=1$ for each decoder column via normalization during training.
\end{enumerate}
A unit-norm constraint will neutralize L2 shrinkage on the decoder (since $\|W_{\mathrm{dec}}\|_F^2$ becomes constant), so the same $\lambda_{\mathrm{w}}$ may behave differently depending on whether decoder normalization is applied. Motivated by this, we explore these choices in a toy MNIST setting and retain SAEBench defaults when moving to language-model activations. When unit-norm decoder columns are enforced, the weight penalty acts as encoder regularization.

\subsection{Cross-seed feature consistency}

To quantify reproducibility, we followed \citet{Paulo2025SparseFeatures} and measured feature similarity across SAEs trained with different random seeds.
For two trained SAEs $A$ and $B$, let $D_A, D_B \in \mathbb{R}^{m\times d}$ be the decoder feature matrices. We compute the absolute cosine similarity matrix $S = |\mathrm{cos}(D_A, D_B)|$ and use Hungarian matching to obtain a one-to-one pairing.  We report:
\textbf{mean max cosine} (average over features of $\max_j S_{ij}$, symmetrized),
\textbf{fraction paired} (fraction with $\max_j S_{ij} > 0.7$), and
\textbf{shared features} (a stricter criterion requiring encoder and decoder Hungarian matchings to agree \emph{and} both similarities exceed $0.7$).

We call a feature \emph{alive} if its encoder–decoder cosine similarity exceeds 0.1 using this threshold as a pragmatic filter for collapsed or weakly self-consistent latents. This excludes both features whose encoder or decoder norm has collapsed to zero during training and the small number of features with weakly aligned encoder–decoder pairs. On our trained SAEs the encoder–decoder cosine distribution is strongly bimodal (Figure~\ref{fig:lm-cosine}, Appendix~\ref{app:topk_dead}), so results are insensitive to the exact threshold value.

\subsection{Steering and evaluation}

For language-model SAEs, we evaluated \emph{feature steering} by injecting a decoder feature vector into the residual stream during text generation:
$r_t \leftarrow r_t + \alpha \, d_i$,
where $d_i$ is decoder feature $i$ and $\alpha$ is the steering strength.
We used non-deterministic decoding (temperature $= 0.7$, top-$p = 0.9$) with generation-only steering and fixed scaling ($\alpha = 5$). Full hyperparameter details appear in Appendix~\ref{app:steering_hparams}.

For each SAE we sampled 300 features uniformly at random from the alive pool (per §2.3), generated auto-interpretability explanations (LLM-generated descriptions of features, scored by how well a second LLM can use them to predict the feature's activations on example texts)for each, and retained features with valid explanations as steering candidates. Each candidate was steered against 3 prompts. Sampling was performed independently per SAE; because regularized and unregularized models are trained as separate trainers, feature indices are not comparable across SAEs and we report aggregate sample-level statistics rather than paired per-feature comparisons.

An LLM judge (GPT-5.1) scored whether the steered output was differently related to the feature's expected concept than the unsteered output on a 1--5 scale (with 3 being no difference). We counted scores $\ge 4$ as successful steering.

\section{Toy Experiments on MNIST}

\subsection{Weight regularization induces an ``aligned core''}

\begin{figure*}[t]
    \centering
    \includegraphics[width=\textwidth]{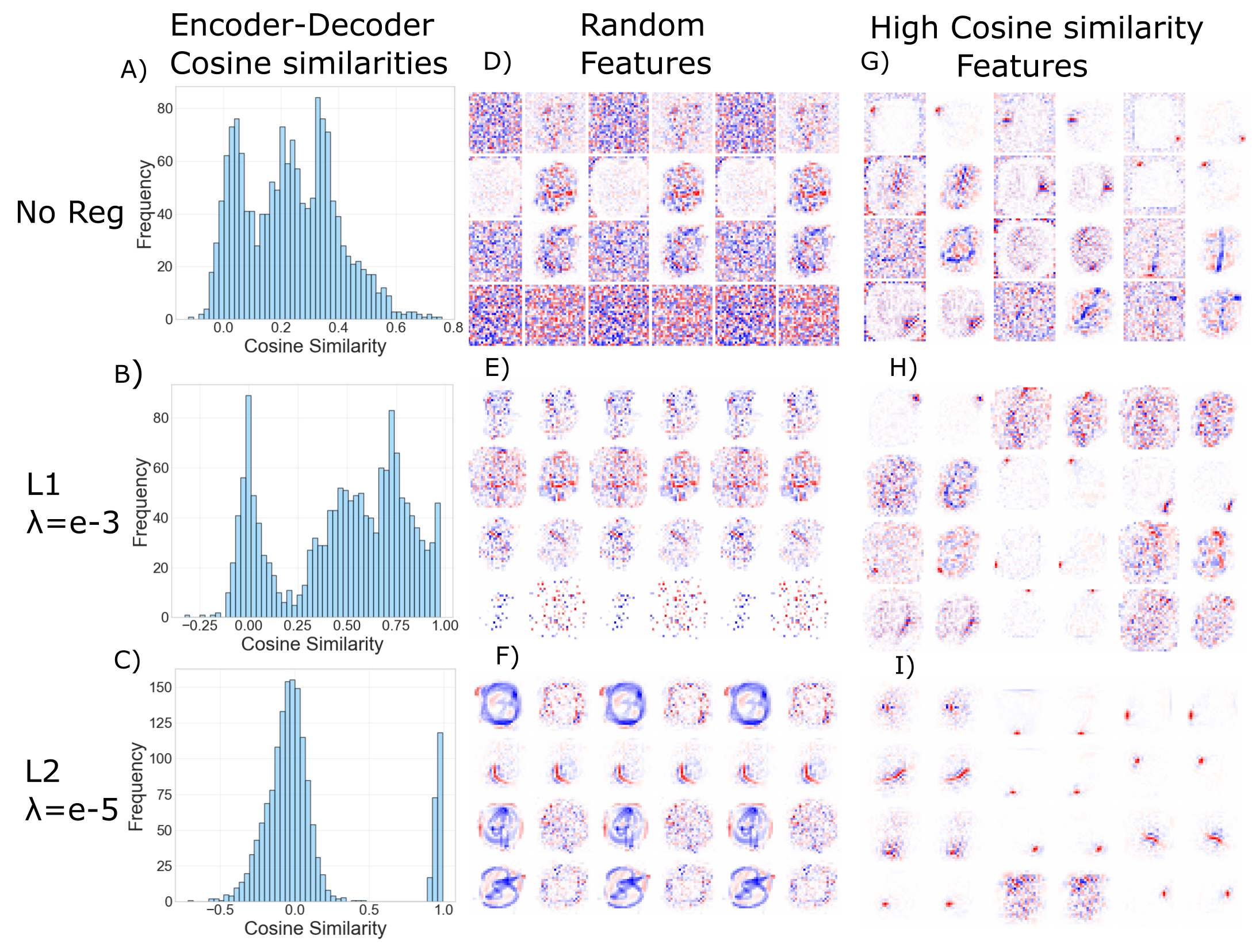}
    \caption{\textbf{MNIST cosine similarity and feature visualizations.} \textbf{Left column (A--C):} Histograms of encoder--decoder cosine similarity for base, L1, and L2 SAEs (1,568 latents). L2 creates a bimodal distribution with a small high-alignment core. \textbf{Right panels (D--I):} Feature pairs showing encoder and decoder as 28$\times$28 heatmaps (blue = negative, red = positive). Left subcolumns show random samples; right subcolumns show high cosine-similarity features. L2's high-alignment features capture clean strokes and curves. The plotted features come from SAEs without decoder norm constraints.}
    \label{fig:mnist}
\end{figure*}

We trained SAEs on flattened MNIST images ($d=784$) with $m=1{,}568$ latents ($2\times$ overcomplete) and an L1 sparsity penalty on activations.
With untied initialization, adding L2 regularization ($\lambda_w=10^{-5}$) produced a bimodal distribution of encoder--decoder cosine similarities and yielded a small set of aligned features that qualitatively correspond to clean strokes and curves (Figure~\ref{fig:mnist}).
A small number of high-cosine similarity latents accounted for most reconstruction capacity (Appendix~\ref{app:mnist_mse}).

\subsection{Regularization improves cross-seed consistency}

To match language-model SAE practices, we repeated MNIST training with tied initialization and a unit-norm decoder constraint to match the design choices present in SAEBench \citep{Karvonen2025SAEBench:Interpretability}.
Table~\ref{tab:mnist_seed_consistency} summarizes cross-seed consistency across three random seeds.

\begin{figure*}[t]
    \centering
    \includegraphics[width=\textwidth]{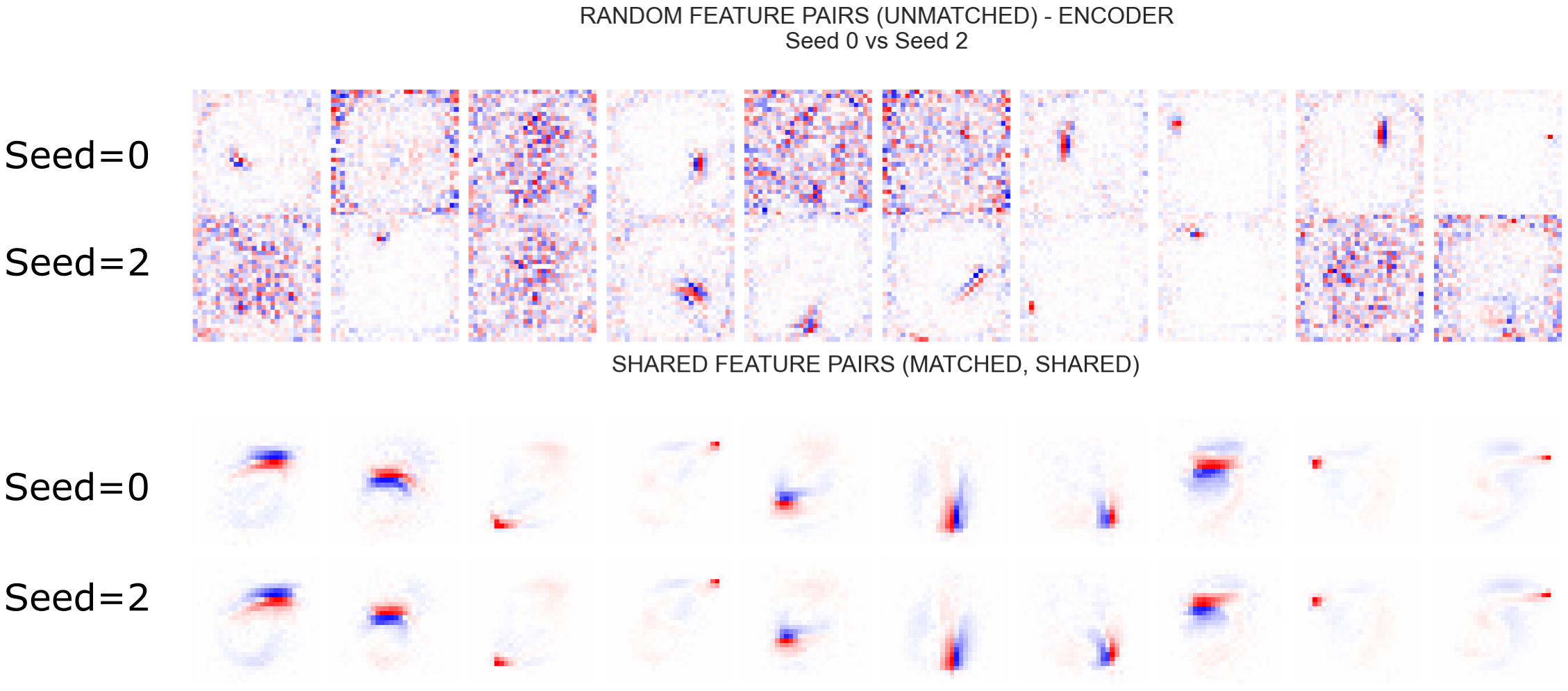}
    \caption{\textbf{MNIST shared vs.\ random feature visualization.} Encoder weights as 28$\times$28 heatmaps (blue = negative, red = positive). Top two rows: random features; bottom two rows: features shared between SAEs with seed 0 and 2. Shared features capture clean strokes and curves; random features appear noisy. Features shown are from the SAE with tied weights and constrained decoder but no weight penalty.}
    \label{fig:mnist-shared-random-encoders}
\end{figure*}

Without decoder constraints and tied initialization, features are mostly not reproducible by the strict shared-feature criterion ($\approx 0\%$ shared). With tied initialization and unit-norm decoders, adding weight regularization substantially increases the fraction of shared features. L2 yields the highest shared-feature fraction for alive features ($22.5\%$ vs.\ $1.9\%$ without regularization). We also observed that shared features appear qualitatively more interpretable than random features (Figure~\ref{fig:mnist-shared-random-encoders}). Based on these results, we proceeded with SAEBench defaults for Pythia experiments.

\begin{table*}[t]
\centering
\small
\begin{tabular}{llccccc}
\toprule
Setting & Weight penalty & Mean max cos & Frac paired ($>0.7$) & Frac shared (All) & Frac shared (Alive) \\
\midrule
Untied init & none & 0.2400 & 3.57\% & 0.00\% & 0.00\% \\
Tied init (constr.\ Dec) & none & 0.4783 & 48.28\% & 1.74\% & 1.74\% \\
Untied init & L1 ($\lambda_w=10^{-3}$) & 0.3260 & 4.66\% & 3.21\% & 0.60\% \\
Tied init (constr.\ Dec) & L1 ($\lambda_w=10^{-3}$) & 0.5486 & 51.62\% & 14.07\% & 14.10\% \\
Untied init & L2 ($\lambda_w=10^{-5}$) & 0.5061 & 7.38\% & 4.91\% & 2.90\% \\
Tied init (constr.\ Dec) & L2 ($\lambda_w=10^{-5}$) & 0.3858 & 42.88\% & 12.54\% & 22.50\% \\
\bottomrule
\end{tabular}
\caption{MNIST cross-seed feature consistency (3 seeds) with unit-norm decoder columns. Weight regularization interacts strongly with decoder constraints and tied initialization; L2 regularization increases the fraction of strictly shared features by an order of magnitude.}
\label{tab:mnist_seed_consistency}
\end{table*}

\section{Language Model Experiments}

\subsection{Setup}

We evaluated weight regularization on SAEs trained on layer-3 residual stream activations from Pythia-70M-deduped using SAEBench \citep{Karvonen2025SAEBench:Interpretability}. We tested TopK \citep{Gao2024ScalingAutoencoders}, BatchTopK \citep{Bussmann2024BatchTopKAutoencoders}, and Matryoshka \citep{Bussmann2025LearningAutoencoders} architectures with sweeps over sparsity ($k \in \{40, 80, 150, 320\}$) and regularization strengths ($\lambda \in \{0, 10^{-12}, 10^{-10}, 10^{-7}, 10^{-5}, 10^{-4}\}$), keeping tied initialization and unit-norm decoder columns throughout.

We evaluate on sparse probing (SP), spurious correlation removal (SCR@10), targeted probe perturbation (TPP@10), and CE loss recovery for all architectures. We then measure cross-seed feature sharedness for the TopK architecture, and proceeded with automated interpretability and steering experiments.

\subsection{SAEBench performance and cosine-similarity structure}

\begin{figure*}[t]
    \centering
    \includegraphics[width=\textwidth]{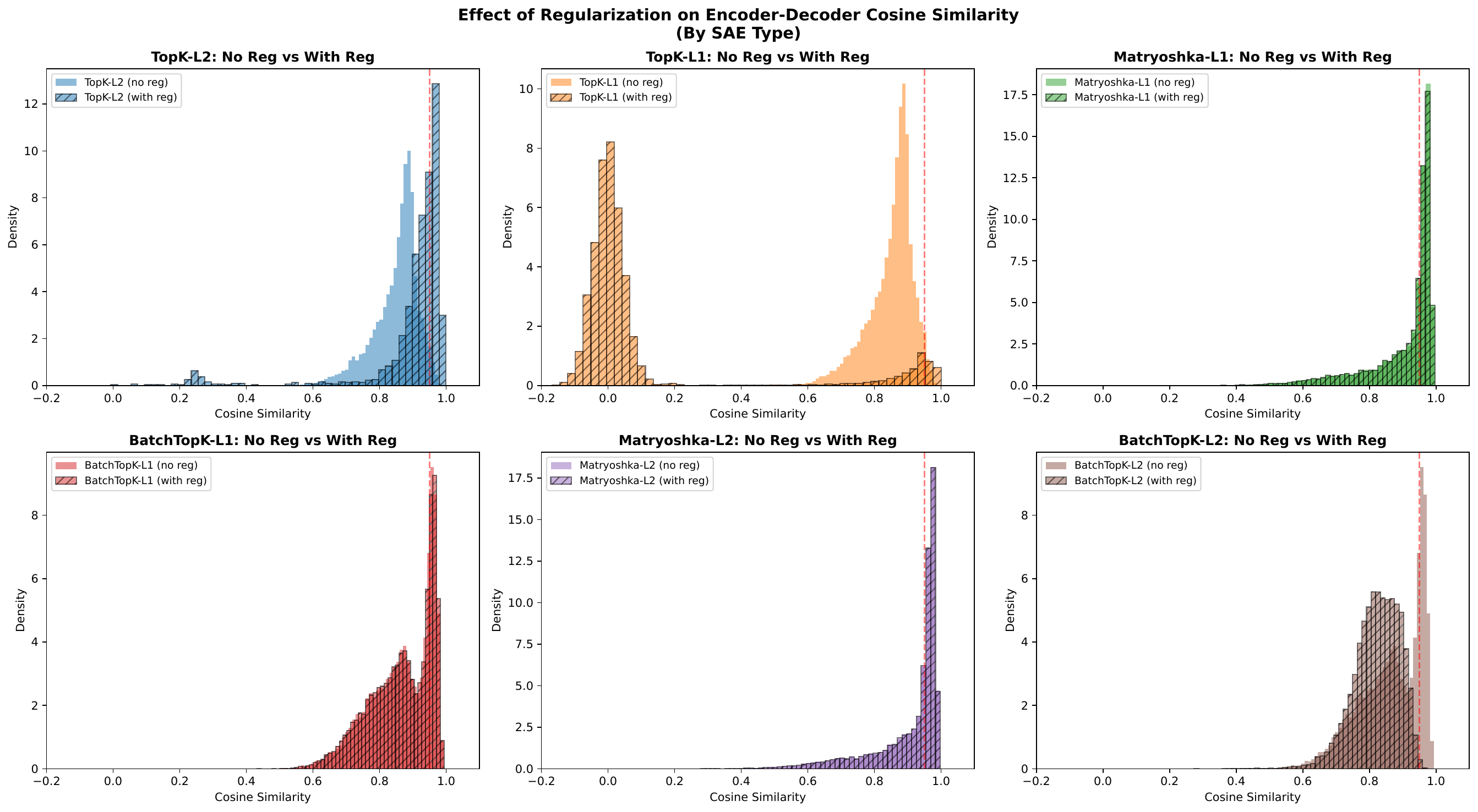}
    \caption{\textbf{Encoder--decoder cosine similarity distributions} for regularized (``reg'') vs.\ unregularized (``no reg'') SAEs across architectures.Features with zero encoder or decoder norm are excluded; for TopK-L2, the majority of features have collapsed to zero norm during training (see Appendix~\ref{app:topk_dead}). Distributions are of the Pareto-best SAE for each architecture and the unregularized SAE with corresponding $k$.}
    \label{fig:lm-cosine}
\end{figure*}

Across architectures, weight regularization frequently appears in Pareto-optimal configurations under SAEBench metrics (Appendix~\ref{app:saebench_pareto}).
As in MNIST, TopK models with L2 regularization produce a high-cosine cluster of aligned features alongside many dead latents (Figure~\ref{fig:lm-cosine} and Appendix~\ref{app:topk_dead}). This effect is architecture-dependent: BatchTopK with L2 regularization instead shows a general shift toward lower cosine similarities without the bimodal structure. In TopK models, L2 regularization is particularly aggressive, killing off the majority of features during training (see Appendix~\ref{app:topk_dead}).

\subsection{L2 regularization increases cross-seed feature sharedness}

\begin{figure*}[t]
  \centering
  \includegraphics[width=\linewidth]{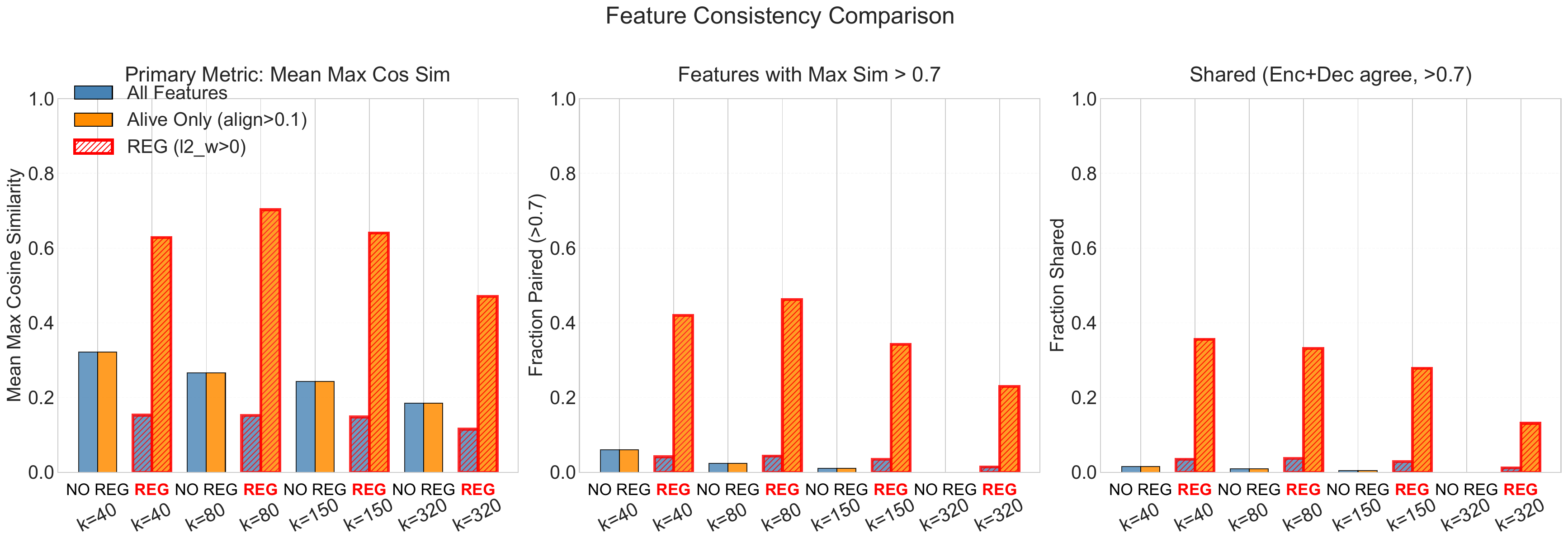}
  \caption{Pythia-70M TopK: cross-seed feature consistency metrics across sparsity levels ($k$) for unregularized (blue) and L2-regularized (orange) SAEs. L2 weight regularization substantially increases sharedness, particularly among alive features.}
  \label{fig:lm_sharedness}
\end{figure*}

We computed cross-seed feature reproducibility across three random seeds for TopK SAEs ($k \in \{40, 80, 150, 320\}$), comparing $\lambda \in \{0, 10^{-4}\}$.
Adding an L2 weight penalty produced a large and consistent improvement across all sparsity levels (Figure~\ref{fig:lm_sharedness}): the fraction of strictly shared features among alive features increases over tenfold (from $<2\%$ to ${\sim}35\%$), mean max cosine similarity among alive features roughly doubles (from ${\le}0.32$ to ${\sim}0.7$), and the fraction of features with cosine similarity $> 0.7$ in decoder increases almost fivefold (from $<10\%$ to ${\sim}50\%$).

\subsection{Regularization improves steering and strengthens the auto-interp--steering link}

\begin{figure*}[t]
  \centering
  \includegraphics[width=0.7\textwidth]{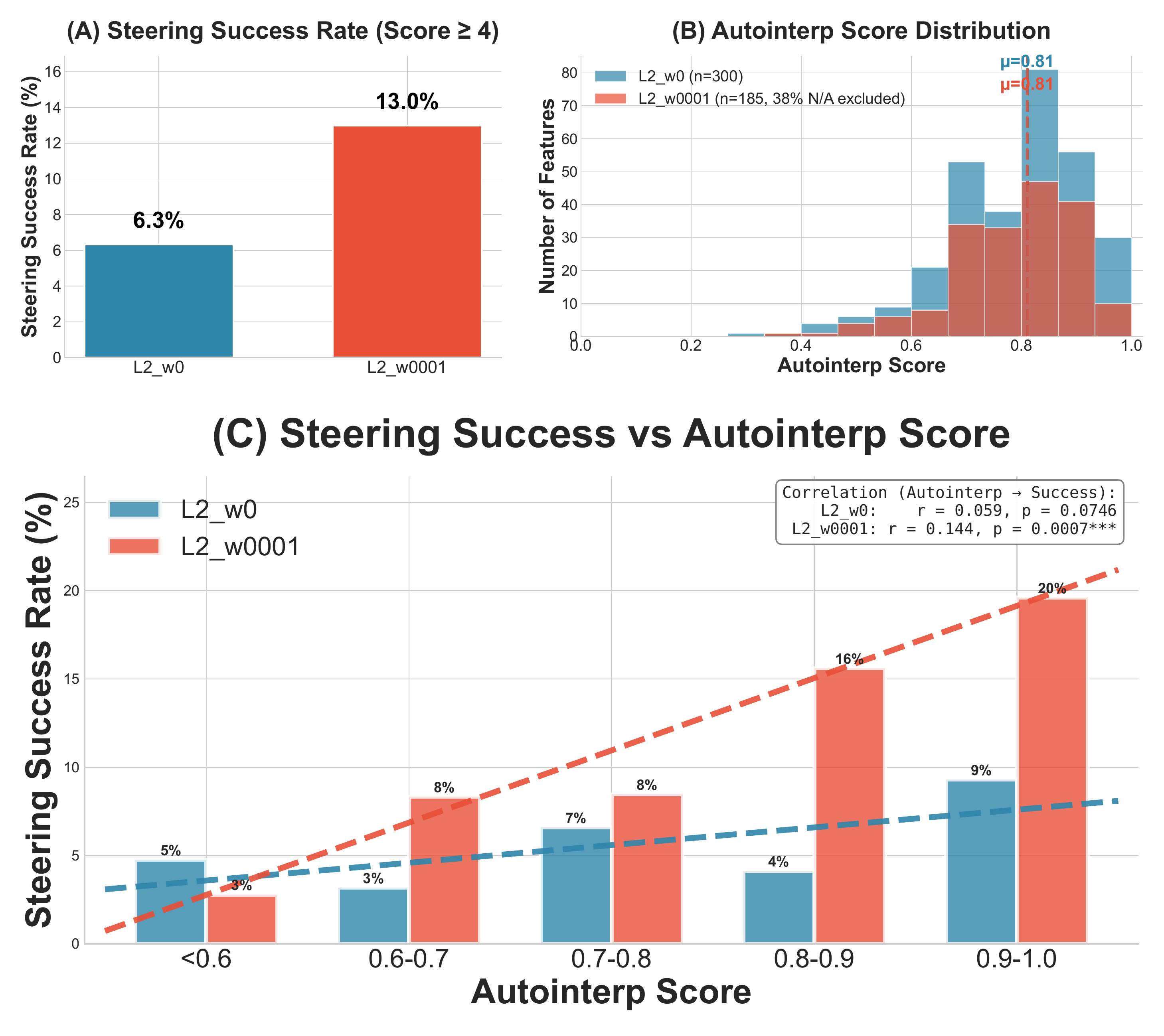}
  \caption{Pythia-70M TopK $k{=}40$: \textbf{(A)} Steering success rate (LLM judge score $\ge 4$); L2 regularization roughly doubles the rate. \textbf{(B)} Auto-interpretability score distributions remain similar across conditions. \textbf{(C)} Spearman correlation between auto-interpretability and steering success; regularization strengthens the link.}
  \label{fig:lm_steering_success}
\end{figure*}

For TopK SAEs at $k{=}40$, L2 weight regularization increased steering success under our sampling protocol. We independently sampled 300 alive features from each SAE, drawn from alive pools of 16{,}384 features for the unregularized model and 1{,}699 features for the L2-regularized model, and steered each feature against 3 prompts. All 300 unregularized features received valid auto-interpretability explanations (900 samples), compared with 185/300 L2 features (555 samples). Conditional on valid explanations, sample-level success rates (judge score $\ge 4$) rose from 6.3\% (57/900) to 13.0\% (72/555) with L2 ($\chi^2 = 17.92$, $p < 10^{-4}$). Thus, the steering improvement should be read as conditional on the explained surviving features; the 115 invalid L2 explanations are themselves informative, indicating a weakly structured tier of alive but hard-to-explain features. Conditional on a valid explanation, the auto-interpretability score distribution is preserved across conditions (Figure~\ref{fig:lm_steering_success}B).

The relationship between auto-interpretability and steering success also strengthens with regularization. The Spearman correlation between auto-interpretability score and per-feature steering success is weak without regularization ($r = 0.060$, $p = 0.075$) but becomes significant with L2 regularization ($r = 0.144$, $p = 7 \times 10^{-4}$). The corresponding slope of success probability against auto-interpretability score increases from $0.11$ (SE $0.06$) without regularization to $0.40$ (SE $0.12$) with L2, a significant change ($Z = 2.21$, $p = 0.027$), indicating that auto-interpretability becomes a meaningfully stronger predictor of steering success under regularization. The distribution of auto-interpretability scores themselves remains similar across conditions (Figure~\ref{fig:lm_steering_success}), suggesting that regularization does not simply improve interpretability scores but instead better aligns what a feature \emph{means} with what it \emph{does}.

\begin{figure*}[t]
\centering
\includegraphics[width=0.9\textwidth]{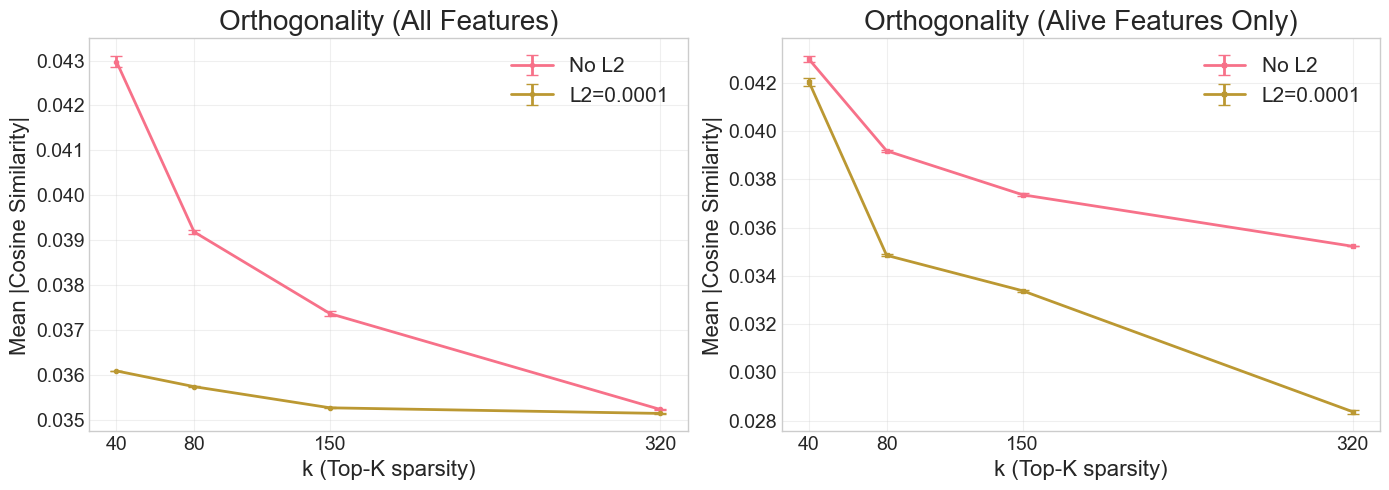}
\caption{Decoder pairwise cosine similarity ($c_\text{dec}$) across TopK sparsity levels 
(mean across 3 seeds; across-seed SDs are $\leq 2 \times 10^{-4}$ and barely visible at this scale). \textbf{Left:} computed over all features; L2 reduces $c_\text{dec}$, 
partly driven by dead features. \textbf{Right:} computed over alive features only 
(encoder--decoder alignment $> 0.1$). At $k{=}40$, alive-feature orthogonality is nearly 
identical; at higher $k$, surviving L2 features are \emph{more} orthogonal than the full 
unregularized dictionary.}
\label{fig:decoder-orthogonality}
\end{figure*}

\paragraph{Decoder orthogonality.}
To test whether L2 regularization encourages more orthogonal decoder columns, we computed the mean absolute pairwise cosine similarity of the decoder matrix ($c_\text{dec}$; \citep{Chanin2025SparseAutoencoders}) for each TopK configuration across 3 seeds (Figure~\ref{fig:decoder-orthogonality}). All values are robust across seeds (across-seed SD $\leq 2\times10^{-4}$, much smaller than the differences discussed below). For context, random unit vectors in $\mathbb{R}^{512}$ have expected mean absolute cosine $\sqrt{2/(\pi d)} \approx 0.035$.
Over all features, L2 consistently lowers $c_\text{dec}$, with the largest reduction at $k{=}40$ ($0.043 \to 0.036$). Restricting to alive features reveals a more nuanced picture: at $k{=}40$ the values are nearly identical ($0.043$ vs.\ $0.042$), while at $k \geq 80$ the surviving L2 features become progressively more orthogonal than the full unregularized dictionary (e.g., $0.028$ vs.\ $0.035$ at $k{=}320$). Notably, the unregularized dictionary's alive-feature $c_\text{dec}$ approaches the random-vector baseline as $k$ increases, while L2 alive features cross \emph{below} the baseline from $k \geq 80$ onward, indicating that regularization actively pushes the dictionary toward greater-than-random orthogonality.

\section{Discussion}

Adding L2 weight regularization to SAE training increases cross-seed feature reproducibility and improves steering success, but the majority of latents collapse to zero (Appendix~\ref{app:topk_dead}). Concretely, the TopK-L2 model retains over ~1,500 alive features which is still ~3× the Pythia-70M residual stream dimension of 512. The regularized model thus remains a standard SAE in both defining respects: sparse activations (fixed by
k) and an overcomplete latent dictionary, just less aggressively so than the nominal 32× expansion. This raises a natural question: is L2 performing implicit dictionary-size selection, and would an unregularized SAE with ~1,500 latents learn similar features? We suspect the answer is largely yes and that much of L2's benefit stems from finding an appropriate effective dictionary size, a choice that currently requires manual sweeps. However, L2 might shape which features survive in ways a smaller unregularized dictionary would not replicate. A direct comparison was outside the scope of this exploration and we leave it to future work. Framed this way, the collapse is not necessarily only a failure mode: even if L2's effect were largely capacity selection, an automatic pressure toward a smaller effective dictionary could still be useful, provided it does not remove features needed for downstream applications. Even without that comparison, two observations support the view that the surviving features are meaningful rather than arbitrary. First, the cross-seed consistency results (Figure~\ref{fig:lm_sharedness}) indicate that independent optimization runs converge on overlapping subsets of surviving features. Second, \citet{Martin-Linares2025Attribution-GuidedAutoencoders} arrive at a structurally similar outcome through an entirely different mechanism: their attribution-guided distillation converges on a core of 197 features (out of 65k) that persist across retraining cycles. That both L2 weight decay and iterative attribution-based pruning dramatically reduce the effective dictionary to a compact subset suggests that standard SAE dictionaries contain substantial redundancy, and that multiple forms of complexity control converge on similar effective sizes.

\paragraph{Interaction with SAE design choices.}
Regularization does not act in isolation. In MNIST, the combination of tied initialization, unit-norm decoder columns, and L2 yielded the highest fraction of alive features shared, an order of magnitude larger than regularization or decoder constraints alone (Table~\ref{tab:mnist_seed_consistency}). The response is also architecture-dependent: TopK models with L2 produce a bimodal cosine-similarity distribution, while BatchTopK shows a general shift without bimodal structure (Figure~\ref{fig:lm-cosine}). These interactions show that the effect of weight regularization is shaped by the full training recipe.

\paragraph{Steering and the interp--steering gap.}
The roughly doubled steering success rate is practically relevant. We measured steering only at $k{=}40$, where the regularised model had the highest fraction of shared features across seeds, making it the most natural test case; the discussion below is therefore directly grounded only at this sparsity level. We initially hypothesised that L2 produces more orthogonal decoder columns, and Figure~\ref{fig:decoder-orthogonality} shows this holds only partially. At $k{=}40$, regularised and unregularised SAEs have nearly identical alive-feature orthogonality, so the steering improvement we observe is better attributed to dictionary pruning: by collapsing $\sim$90\% of latents, L2 removes weakly-used or redundant directions that would otherwise introduce off-target interference when steering. At higher $k$, the surviving L2 features are more orthogonal than the full unregularised dictionary (Figure~\ref{fig:decoder-orthogonality}, right), suggesting that regularisation can also produce a more disentangled basis beyond pruning alone. Whether this orthogonality gain translates into further steering improvements at higher $k$ -- i.e.\ whether the mechanism behind improved steering genuinely shifts with sparsity -- is an open question, since our steering experiments are limited to $k{=}40$ and the orthogonality gains are small in absolute terms. Independently, the strengthened correlation between auto-interpretability and steering success (Spearman $r$: $0.060 \to 0.144$) suggests that regularisation may partially narrow the gap between what a feature \emph{means} and what it \emph{does}.

A complementary approach is to modify the SAE objective so that learned features are explicitly \emph{functionally important}. \citet{Braun2024IdentifyingLearning} propose end-to-end sparse dictionary learning, training an SAE to minimize divergence between original and reconstructed model outputs. Our findings show that even a simple L2 weight penalty can partially align textual explanations with controllability; combining weight regularization with end-to-end objectives might be a promising direction.

\paragraph{Connections to prior work.}
The dead-feature phenomenon can also be reinterpreted through the lens of the Minimum Description Length principle \citep{Ayonrinde2024InterpretabilityMDL-SAEs}:L2 may eliminate features whose marginal reconstruction contribution does not justify their coding cost. Under this view, L2 regularization and attribution-guided selection \citep{Martin-Linares2025Attribution-GuidedAutoencoders} act as complementary forms of complexity control — one continuous during optimization, the other discrete between training cycles — and their convergence on compact feature sets is what an MDL perspective would predict.

\paragraph{Limitations and future work.}
All language-model experiments use Pythia-70M-deduped; scaling behavior is unknown. Cross-seed and steering analyses focus on TopK, though Figures~\ref{fig:lm-cosine} and Appendix~\ref{app:saebench_pareto} suggest architecture-dependent responses. Steering evaluation relies on a single LLM judge, and the high dead-feature rate limits applications requiring comprehensive coverage. Key next steps include directly testing the implicit-capacity-selection hypothesis by comparing regularized SAEs against smaller unregularized dictionaries matched on alive-feature count — a comparison that would disentangle how much of L2's benefit is explained by dictionary size alone versus additional shaping of which features are learned — along with developing diagnostics that distinguish useful from pathological feature death, exploring regularization annealing schedules, and combining weight regularization with post-hoc feature selection methods such as attribution-guided distillation \citep{Martin-Linares2025Attribution-GuidedAutoencoders}.

\section{Conclusion}

Weight regularization is a simple modification that meaningfully affects SAE behavior.
In MNIST, it yields a small aligned core of clean features and under standard training constraints substantially increases cross-seed feature sharedness.
In Pythia-70M TopK SAEs, L2 weight regularization increases the fraction of shared features across seeds and improves steering success among features with valid explanations, but also collapses many latents and increases the invalid-explanation rate. We therefore interpret L2 primarily as a simple form of complexity control or implicit effective-dictionary-size selection. Regularization also modestly strengthens the relationship between auto-interpretability scores and steering success, suggesting a possible pathway toward more reliable and functionally meaningful SAE features.

\section*{Impact Statement}

This paper presents work whose goal is to advance the field of mechanistic interpretability. By improving the stability and reliability of sparse autoencoders, this work may contribute to better tools for understanding and monitoring neural network behavior, with potential benefits for AI safety and AI for Scientific discovery. There are many potential societal consequences of our work, none which we feel must be specifically highlighted here.

\section*{Acknowledgements}
We thank Peter Minary, Craig MacLean, and Adam Winnifrith for helpful
feedback and discussion.
PJ is supported by the Biotechnology and Biological Sciences Research
Council (UKRI-BBSRC) grant BB/T008784/1. OMC acknowledges funding from a
New College Todd-Bird Junior Research Fellowship and MRC Fellowship
MR/Y010078/1. 

\bibliography{references}
\bibliographystyle{icml2026}

\appendix
\onecolumn

\section{Steering Hyperparameters}
\label{app:steering_hparams}

We ran a grid search over steering hyperparameters including: (i) deterministic vs.\ sampling decoding, (ii) applying steering during generation only vs.\ all forward passes, and (iii) different strength parameterizations (fixed, residual-RMS scaled, target pre-activation delta).
Across both regularized and unregularized TopK models, residual-RMS scaling gave the best average LLM-judge score, while fixed-strength steering often gave higher ``high-change'' rates (fraction of samples with judge score $\ge 4$).
The configuration used for the main analyses was non-deterministic decoding (temperature $= 0.7$, top-$p = 0.9$) with generation-only steering and fixed scaling ($\alpha = 5$), as it gave the best results for both regularized and unregularized SAEs.

\section{MNIST Reconstruction MSE Ablation}
\label{app:mnist_mse}

\begin{figure}[h]
    \centering
    \includegraphics[width=0.7\textwidth]{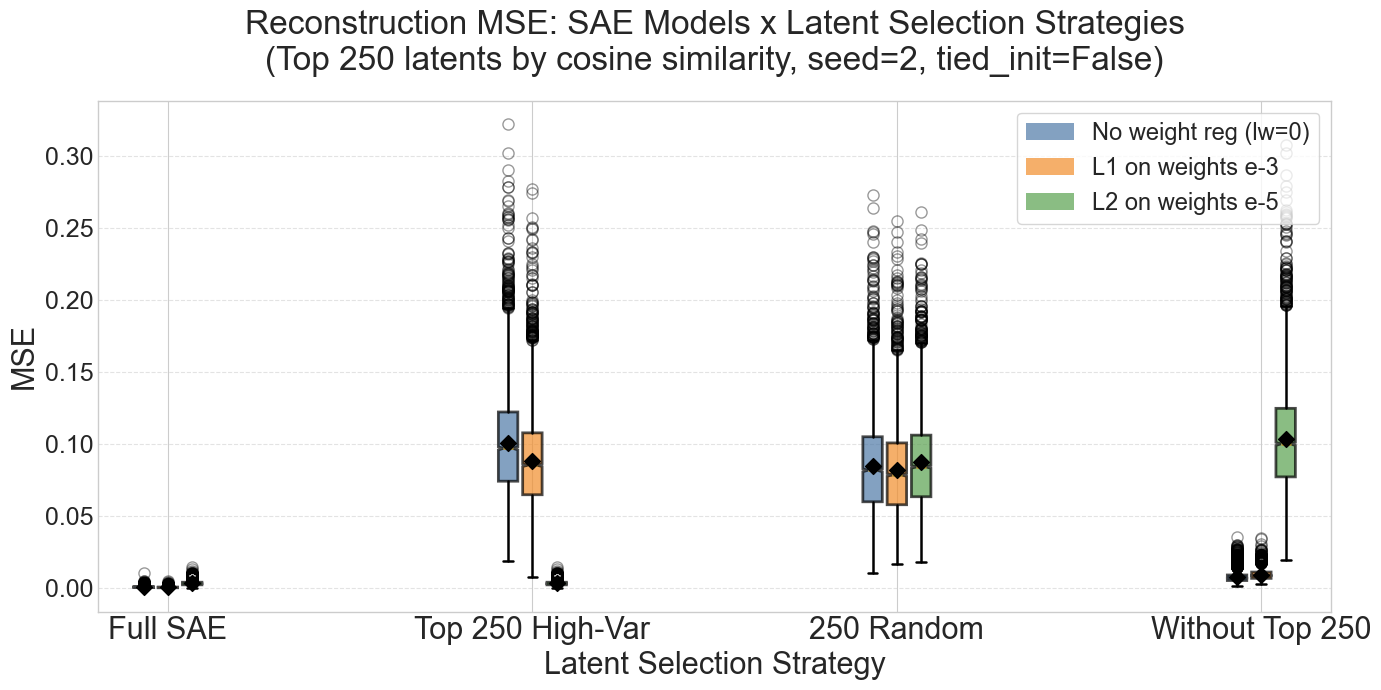}
    \caption{\textbf{MNIST reconstruction MSE.} Boxplots measured on the first 250 MNIST examples for three SAE variants (base, L1, L2) under four latent selection strategies: full SAE (all latents active), top-250 high-cosine-similarity latents, 250 random latents, and all latents except the top-250. A small number of high-cosine similarity latents accounts for most reconstruction capacity in L2-regularized SAEs.}
    \label{fig:mnist-mse}
\end{figure}

\section{TopK-L2 Cosine Similarity Including Dead Features}
\label{app:topk_dead}

\begin{figure}[h]
    \centering
    \includegraphics[width=\textwidth]{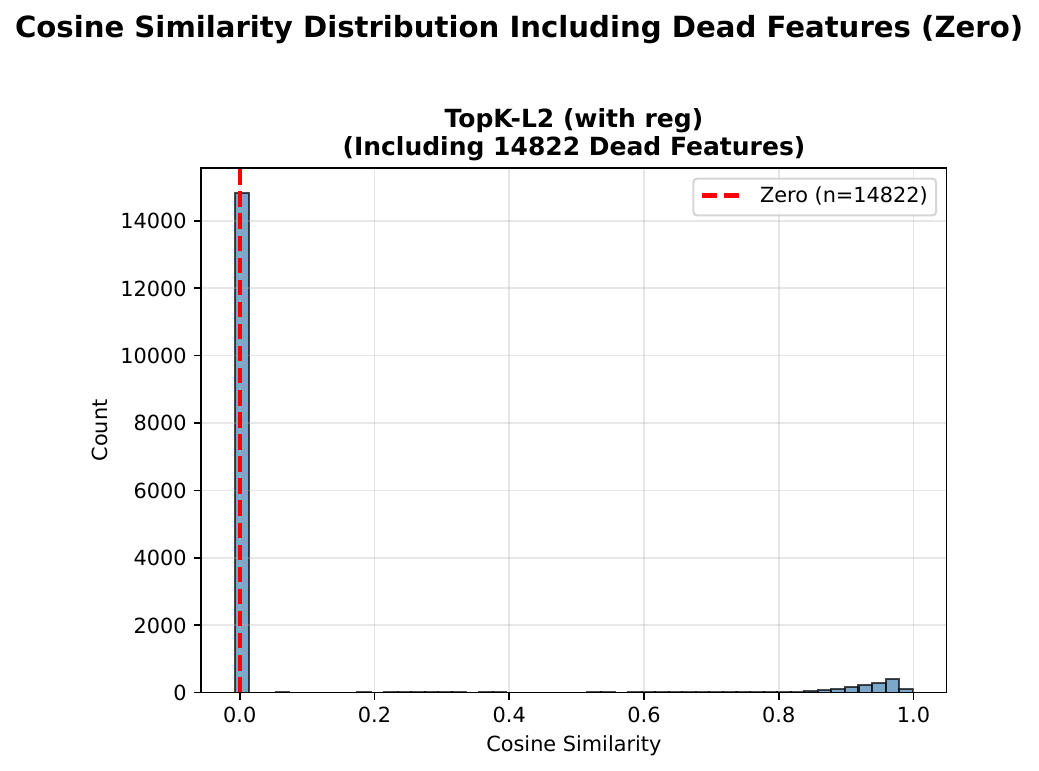}
    \caption{\textbf{TopK cosine similarity including dead features.} Encoder--decoder cosine similarity distributions for regularized vs.\ unregularized TopK SAEs, retaining features where the encoder or decoder norm is zero. The majority of L2-regularized features collapse to zero norm during training.}
    \label{fig:lm-cosine-topkl2}
\end{figure}

\FloatBarrier

\section{SAEBench Pareto Analysis}
\label{app:saebench_pareto}

\begin{figure}[h]
    \centering
    \includegraphics[width=\textwidth]{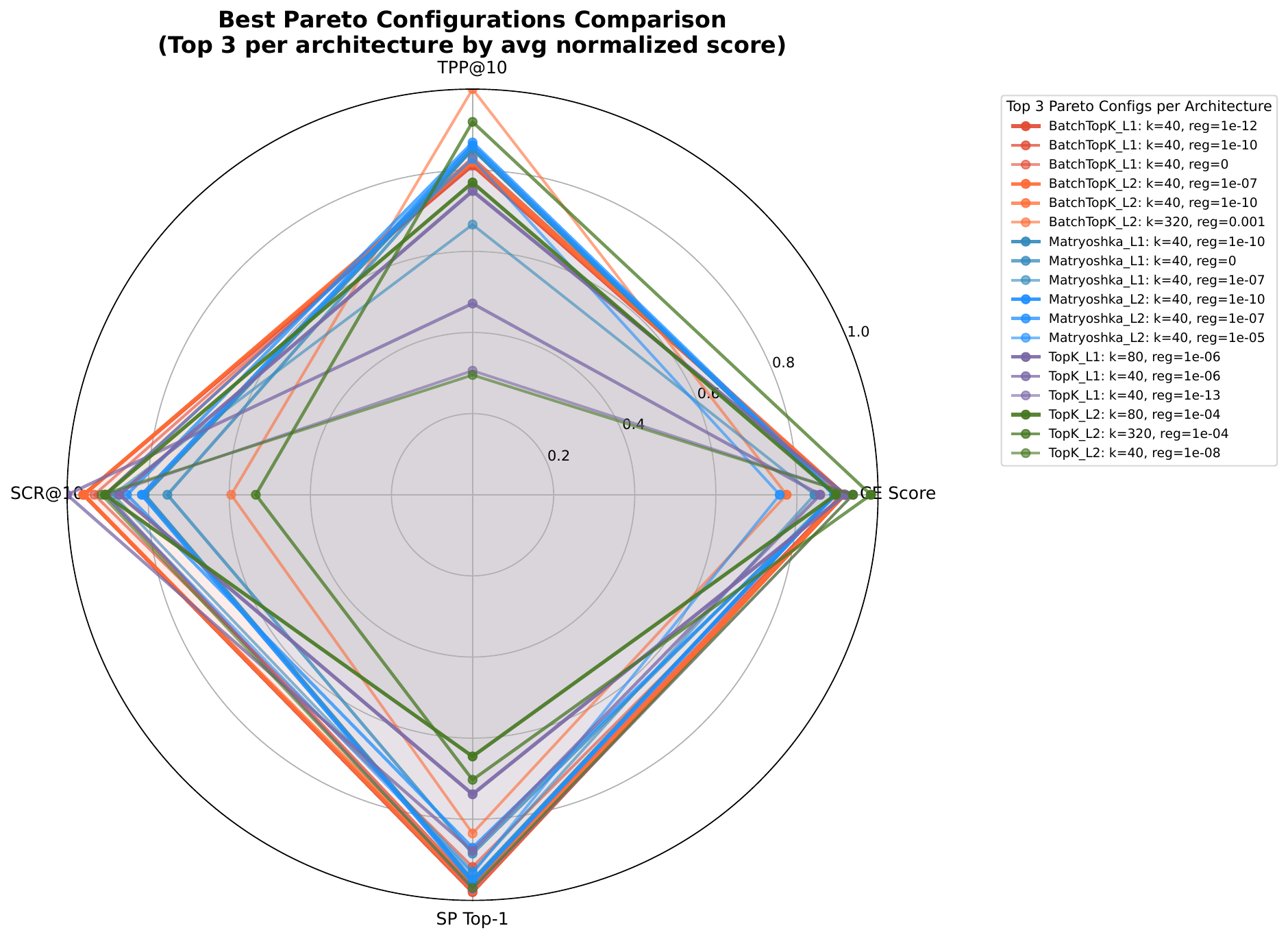}
    \caption{\textbf{SAEBench Pareto analysis.} Aggregated top-3 Pareto configurations per architecture. Regularized models frequently appear on the Pareto frontier. Radar plots showing normalized SAEBench metrics. Red = Pareto-optimal; gray = dominated. Axes: TPP@10, CE score, SCR@10, SP Top-1 (higher = better).}
    \label{fig:saebench_pareto}
\end{figure}

\end{document}